\pdfoutput=1

\documentclass[11pt]{article}

\usepackage[final]{acl}

\usepackage{times} 
\usepackage{latexsym}
\usepackage{graphicx}
\usepackage{multirow}
\usepackage{arydshln}
\usepackage[capitalise, noabbrev]{cleveref}

\usepackage[T1]{fontenc}

\usepackage[utf8]{inputenc}
\usepackage{amssymb} 
\usepackage{booktabs}
\usepackage{microtype}
\usepackage{inconsolata}
\usepackage{xspace}
\usepackage[]{todonotes}
\usepackage[normalem]{ulem} 
\usepackage{xcolor}

\def\change#1{#1}
\def\DatasetName{\emph{MockConf}}
\def\ToolName{\emph{InterAlign}}
\def\SystemName{BA}
\def\SystemSubName{BA+sub}
\def\SystemLabName{BA+sub+lab}
\def\BaselineName{Baseline}

\title{\DatasetName{}: A Student Interpretation Dataset: \\ Analysis, Word- and Span-level Alignment and Baselines}

\author{Dávid Javorský$^1$ \and Ondřej Bojar$^1$ \and François Yvon$^2$
  \\ \\
  $^{1}$Charles University, Faculty of Mathematics and Physics, Prague, Czechia\\
  $^2$Sorbonne Université, CNRS, ISIR, Paris, France\\
  \texttt{\{javorsky,bojar\}@ufal.mff.cuni.cz}\quad \texttt{francois.yvon@cnrs.fr}}

\begin{document}
\maketitle

\begin{abstract}
In simultaneous interpreting, an interpreter renders a source speech into another language with a very short lag, much sooner than sentences are finished. In order to understand and later reproduce this dynamic and complex task automatically, we need dedicated datasets and tools for analysis, monitoring, and evaluation, such as parallel speech corpora, and tools for their automatic annotation. Existing parallel corpora of translated texts and associated alignment algorithms hardly fill this gap, as they fail to model long-range interactions between speech segments or specific types of divergences (e.g., shortening, simplification, functional generalization) between the original and interpreted speeches. In this work, we introduce \DatasetName, a student interpreting dataset that was collected from Mock Conferences run as part of the students' curriculum. This dataset contains 7~hours of recordings in 5~European languages, transcribed and aligned at the level of spans and words. We further implement and release \ToolName, a modern web-based annotation tool for parallel word and span annotations on long inputs, suitable for aligning simultaneous interpreting. We propose metrics for the evaluation and a baseline for automatic alignment. Dataset and tools are released to the community.
\end{abstract}

\section{Introduction}
Recent advances in speech and translation technologies offer new perspectives for the study of multilingual speech processing, a field that has its origins several decades ago \citep{waibel04_interspeech}. This includes, for instance, the translation of speech transcripts for videos, to be used as captions in a video player, or the automatic generation of full-fledged subtitles for movies or TV shows. These processes have already been studied, and resources are available for a variety of genres and languages, enabling the development of automatic end-to-end subtitling systems \cite{rousseau-etal-2012-ted, cettolo-etal-2012-wit3, lison-tiedemann-2016-opensubtitles2016, pryzant-etal-2018-jesc, di-gangi-etal-2019-must, karakanta-etal-2020-must}. Other speech translation tasks have been considered, involving an increased level of interactivity, such as multilingual information systems \citep{van-den-heuvel-etal-2006-tc}, or translation tools for mediated conversations in various contexts, e.g.\ interactions between patients and doctors \citep{rayner2000spoken, ji2023translation} or military applications \citep{stallard2011bbn}. For these tasks, translations can happen in turns and the focus is often on the informational adequacy of the translated content. 

\begin{figure}[t]
    \centering
    \includegraphics[width=\linewidth]{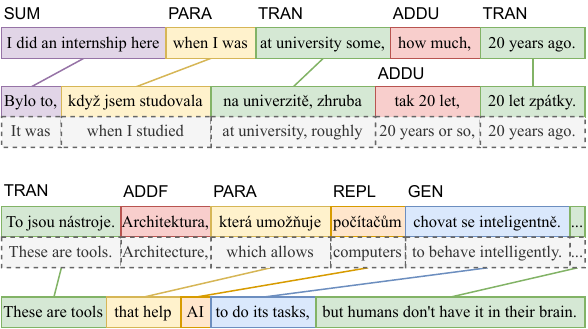}
    \caption{Examples of span-level annotation from our dataset. The first and second rows display transcripts of the original speech and its interpretation. The gray dashed row is the gloss of the Czech part. Span labels are displayed above the corresponding spans, see \cref{tab:labels} for a description of labels.}
    \label{fig:annotation_example}
\end{figure}

In this study, we focus on another type of multilingual task: simultaneous interpreting.\footnote{Defined by \citet{Diriker-2015-simultaneous} as: “Broadly speaking, simultaneous interpreting (SI) is the mode of interpreting in which the interpreter renders the speech as it is being delivered by a speaker into another language with a minimal TIME LAG of a few seconds.”} This mode of interpretation typically occurs in international conferences, where a presenter’s speech is immediately rendered into a foreign language. Simultaneous interpreting has been an active area of research, particularly thanks to resources derived from institutions such as the European Parliament \cite{machacek21_interspeech} and, more recently, ACL conferences \citep{agrawal-etal-2023-findings}.

\change{Building on this research, we introduce \DatasetName{}, a dataset centered on Czech, comprising simultaneous interpreting data with human-annotated transcriptions at both the span and word levels. The dataset creation process involves several key steps: First, we obtain a faithful transcription of human simultaneous interpretings that were collected from Mock Conferences run as part of the student interpreters curriculum. These data was then manually aligned and annotated at the word and span level using \ToolName{}, a dedicated tool designed to facilitate the annotation at the span and word levels. Some example annotations are shown in \cref{fig:annotation_example}. Additionally, we propose a new automatic alignment task that aims to reproduce these manual alignments. In our experiments, we establish baselines and discuss the challenges associated with this task.}


\change{\DatasetName{}, serves multiple purposes. First, it offers valuable opportunities for linguistic analyses \citep{doi-etal-2024-word, wein-etal-2024-barriers}, some of which we have already explored. Second, span-level annotations are beneficial for the development and evaluation of automatic alignment tools. Alignments can aid in tasks such as detecting MT hallucinations \citep{pan-etal-2021-contrastive, guerreiro-etal-2023-optimal, dale-etal-2023-detecting} or MQM evaluation using error span classification \citep{burchardt-2013-multidimensional, kocmi-federmann-2023-gemba, li-etal-2025-mqm, lu-etal-2025-mqm}.
\DatasetName{} can also be useful for educational purposes, e.g., to automatically monitor and analyze the productions of student interpreters, or to evaluate human interpreting \citep{stewart-etal-2018-automatic, wein-etal-2024-barriers, makinae2025automaticqualitymetricevaluating}. Finally, the dataset can contribute to the evaluation of automatic simultaneous interpreting systems \citep{wang-etal-2023-japanese}. The \DatasetName{}\footnote{\url{https://github.com/J4VORSKY/MockConf}} dataset with the analysis and baselines, and the \ToolName{}\footnote{\url{https://github.com/J4VORSKY/InterAlign}} annotation tool are publicly released to the community.}

\section{\DatasetName{}: A dataset of simultaneous interpreting \label{sec:datasets}}


\subsection{Recordings and data collection}

The dataset was collected from Mock Conferences that took place as part of the interpreting curriculum at a university. During these conferences, a student plays the role of some celebrity
and prepares a speech on some predefined topic. Students who are enrolled in Master's level studies listen to the speech and interpret it. The interpreters are familiar with the topic and are provided with a short description of the content. The languages covered are Czech, English, French, German, and Spanish and each direct interpreting is always from or into Czech. There are also \emph{relay interpretings}, which are analogous to pivot translations: talks in foreign language are interpreted into Czech, from which they are further interpreted into other languages. All recordings have been automatically transcribed using WhisperX \citep{bain2022whisperx}, then manually revised by native Czech speakers, with sufficient self-reported proficiency in the respective foreign language. Transcribers were asked to capture exactly what was said, even though utterances might contain disfluencies such as hesitations and false starts, or even translation errors. They also labeled spans containing proper names, which we will further use for anonymization purposes. The full transcription guidelines are in \cref{app:transcript_guidelines}.

\begin{table}[t]
    \centering
\scalebox{0.86}{%
\begin{tabular}{l|cc|cc|cc}
& \multicolumn{2}{|c|}{Language}& \multicolumn{2}{c}{Recordings} & \multicolumn{2}{|c}{Token count}\\\hline
split & src & trg & count & duration & \# src & trg \\\hline\hline
\multirow[c]{4}{*}{dev} & cs & de & 2 0 & 00:21:08 & 2377 & 2187 \\
 & cs & en & 0 6 & 01:06:56 & 7876 & 7001 \\
 & cs & es & 0 1 & 00:11:20 & 1370 & 988 \\
 & cs & fr & 1 0 & 00:20:07 & 1922 & 2196 \\
\cline{1-7} \cline{2-7}
all & & & 3 7 & 01:59:31 & 13545 & 12372 \\
\hline\hline
\multirow[c]{8}{*}{test} & cs & de & 1 2 & 00:30:27 & 3211 & 2833 \\
 & cs & en & 0 6 & 01:00:46 & 6819 & 6118 \\
 & cs & es & 0 3 & 00:31:22 & 2873 & 2810 \\
 & cs & fr & 3 0 & 00:29:29 & 3858 & 3789 \\
 & de & cs & 2 0 & 00:21:14 & 2299 & 1840 \\
 & en & cs & 0 5 & 01:02:27 & 9070 & 6395 \\
 & es & cs & 0 2 & 00:19:19 & 2360 & 1837 \\
 & fr & cs & 4 1 & 00:46:12 & 7229 & 4791 \\\hline
 all & & & 10 19 & 05:01:16 & 37719 & 30413 \\
\end{tabular}
}
    \caption{Main statistics of \DatasetName{}. We identify languages with ISO-632-2 codes. The values in the ``count'' cell denote the number of recordings with consent to publish only transcripts or both transcripts and audio, respectively. Tokens are obtained using Moses tokenizer.\protect\footnotemark}
    \label{tab:dataset-description}
\end{table}

\paragraph{Consent to publish}
\footnotetext{\url{https://pypi.org/project/mosestokenizer/}}
We asked each participant for their consent to redistribute their recordings and ended up with around 7 hours of recordings for which we obtained consent from the two participants (speaker and interpreter), which we split into development and test set with a 1:3  ratio. \change{Note that development set is limited to only cs$\rightarrow$xx direction and does not proportionally represent all annotators. We assume that evaluating on such data might lead to a better generalization.}
Participants were allowed to choose between: no consent (excluded from the data), partial consent (to publish the transcripts) and full consent (to publish transcripts and also the voice recordings).
The duration of recordings for which we can publish only the transcripts amounts to 41:15 and 1:36:29 for dev and test sets. Consent to publish also the audio was given for an additional amount of 1:18:16 and 3:24:47 for dev and test set, respectively. Statistics regarding \DatasetName{} are in \cref{tab:dataset-description}; more details for each recording pair can be found in \cref{app:detailed_statistics} and in \cref{app:domains}, where we list the conference main themes. We have also collected an equivalent amount of recordings with consent from only one of the participant students; these are not used in this study \change{and are reserved for the future creation of training data}.

\subsection{\ToolName{}: Our annotation tool}


After transcriptions, a second layer of annotations consists of alignments between the source and target speeches. We perform this alignment for transcripts of complete speeches. Existing tools are designed mainly to align parallel textual corpora of translations, which differ from our transcripts in many ways: for instance, we cannot rely on existing sentence correspondences \citep{zhao-etal-2024-naist}, which is also illustrated in \cref{fig:annotation_example}. We therefore implemented and used our own annotation tool, \ToolName{}, with the main focus on facilitating the annotation process of interpreting spans and word alignments. We discuss existing tools and their limitations in \cref{app:annotation_tools}, as well as the implementation and usage details of \ToolName{}.

\subsection{Annotation guidelines and process}
\label{sec:annotation_process}

\paragraph{Span-level annotation}

The goal of the span-level alignment is to help us monitor and analyze the interpreting process: to separate parts that are adequate and precise \emph{translations} from \emph{reformulations}, where the interpreter needed to compress its translation for the sake of time, and from \emph{errors}. \emph{Reformulations} happen when interpreters are cognitively overloaded or decide that the audience in the target language could be similarly overloaded and adopt strategies such as \emph{generalization}, \emph{summarization}, or \emph{paraphrasing} \cite{al2000use,he-etal-2016-interpretese}. Generally, we define \emph{reformulations} as a less literal version of translations that convey the same meaning in the given context. For errors, we consider the taxonomy of translation departures in simultaneous interpreting designed by \cite{barik1994description} consisting of \emph{omissions}, \emph{additions},\footnote{Additions in the source side correspond to omissions in interpreting and vice versa.} and \emph{replacements}. We further sub-categorize additions and omissions as \emph{factual} or \emph{uninformative}. The difference between them is that \emph{factual} omissions (resp.\ additions) alter the amount of information conveyed, whereas \emph{uninformative} omissions (resp.\ additions) do not. A similar labeling system is used by \citet{doi-etal-2021-large, zhao-etal-2024-naist}. The list of span labels is in \cref{tab:labels}.

\begin{table}[t]
    \centering
    \scalebox{0.91}{%
    \begin{tabular}{l|l|l}
        \multicolumn{3}{c}{Label types} \\\hline
         category & subcategory & label \\\hline\hline
         Translation & - & TRAN \\\hline
         \multirow[c]{3}{*}{Reformulation} & Paraphrase & PARA \\
         & Summariaztion & SUM \\
         & Generalization & GEN \\\hline
         \multirow[c]{2}{*}{Addition} & Factual & ADDF \\
          & Uninformative & ADDU \\\hline
         Replacement & - & REPL \\
    \end{tabular}
    }
    \caption{Label types and their subcategories.}
    \label{tab:labels}
\end{table}

\paragraph{Word-level annotation} For each span-aligned pair, we also annotate word alignments. We forbid word alignment links between different span pairs. We define word alignment as \textit{sure} if the corresponding pair of words is a context-independent translation and as \textit{possible} if the context is needed or a grammatical dependency is required \cite{bojar-prokopova-2006-czech} to understand the correspondence. An example annotation is in \cref{fig:alignment-list} \change{in \cref{app:annotation_tools}.}\\

For this alignment process, we recruited 5 professional translators, all of them native Czech speakers, who were paid 200~CZK per hour. \change{The total cost of annotating the whole dataset was 25~000~CZK.} The annotator guidelines as well as the precise definitions of labels are in \cref{app:guidelines}; the activity of each annotator is in \cref{tab:annotators}.

\begin{table}[t]
    \centering
    \setlength{\tabcolsep}{0.5em}
    \scalebox{0.9}{%
    \begin{tabular}{l|c|cc|cc}
\multicolumn{2}{}{} & \multicolumn{2}{|c}{Development} & \multicolumn{2}{|c}{Test} \\\hline
id & lang. & count & duration & count & duration \\\hline\hline
1 & de & 1 & 00:09:47 & 5 & 00:51:41 \\
2 & en & 5 & 00:55:47 & 3 & 00:30:13 \\
3 & en & 1 & 00:11:09 & 8 & 01:33:00\\
4 & es & 2 & 00:22:41 & 5 & 00:50:41 \\
5 & fr & 1 & 00:20:07 & 8 & 01:15:41\\
\end{tabular}
}
    \caption{Summary of annotators's activity on the development and test sets.}
    \label{tab:annotators}
\end{table}

\section{\DatasetName{}: Properties and analysis \label{sec:analysis}}

\subsection{Annotation differences per annotator}

\begin{figure}
    \centering
    \includegraphics[width=\linewidth]{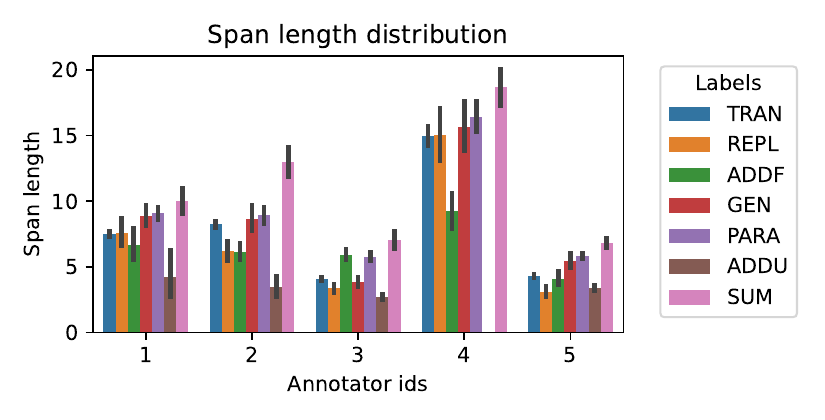}
    \caption{Span length (in tokens) distribution per label and per annotator. The annotators are denoted by their ids which are consistent with \cref{tab:annotators}.
    }
    \label{fig:annotator_span_length}
\end{figure}

\paragraph{Granularity} \cref{fig:annotator_span_length} displays the distribution of span lengths across labels and annotators. The data reveal notable differences in annotator styles, particularly in the lengths of the spans they identify. Annotator~4 consistently reports longer spans --- nearly twice as long as those of other annotators. In contrast, Annotators 3 and 5 tend to annotate much shorter spans. These differences may stem from two potential factors: (1) variability in the annotators' interpretation of the boundary between translation and non-translation, or (2) a lack of adherence to the annotation guidelines.

\change{We believe that the major factor influencing the outputs in \cref{fig:annotator_span_length} is the former. For example, a paraphrase might be labeled as a single span by one annotator, while another might use a more fine-grained approach, resulting in multiple spans. This stems from the fact that, at the token level, distinctions between translations and synonyms / paraphrases can be ambiguous.}

\paragraph{Inter-annotator agreement}

\begin{table}[t]
    \centering
    \setlength{\tabcolsep}{3pt}
    \scalebox{0.91}{%
    \begin{tabular}{cc|cc|cc|cc}
        \multicolumn{4}{c|}{} & \multicolumn{4}{c}{Exact match} \\\hline
        \multicolumn{2}{c|}{segmentation} & \multicolumn{2}{c|}{label} & \multicolumn{2}{c|}{Ann2-Ann3} & \multicolumn{2}{c}{Ann3-Ann2} \\\hline\hline
        src & tgt & src & tgt & w/ & w/o & w/ & w/o \\\hline
        0.56 & 0.57 & 0.41 & 0.25 & 14.85 & 24.26 & 19.87 & 30.46 \\ 
    \end{tabular}
    }
    \caption{Cohen's Kappa for segmentation and label prediction, and the percentage of links the annotators agree upon with the distinction labels vs. no-labels.}
    \label{tab:agreement}
\end{table}

To better understand the differences between annotators, we annotated one recording from the development set twice. The selected recording involves Czech and English and was annotated by two annotators.\footnote{We chose this language pair because it was the only one with two annotators available.} We computed Cohen's Kappa for segmentation (a binary decision regarding span boundaries) and for label agreement, evaluated at the token level (assigning span labels to individual tokens). Additionally, we assessed whether alignment links match, counting both exact matches (corresponding both to similar span boundaries and labels) or a less strict matches (disregarding labels).

The results presented in \cref{tab:agreement} show the following trends: for segmentation, Cohen's Kappa scores are 0.56 and 0.57 for the source and target sides, indicating moderate agreement \citep{landis1977measurement}. For label agreement, the scores are 0.41 and 0.25 for the source and target sides, corresponding to moderate and fair agreement, respectively. The proportions of identical alignment links are 14.85\% (with labels) and 24.26\% (without labels) when using annotator~3 as the reference. In the reverse direction, these proportions increase to 19.87\% and 30.46\%. Upon further inspection, we attribute this discrepancy to the fact that annotator~2 produced fewer alignment links. \change{See \cref{app:annotator_disagreement} for an example of such disagreement.} Overall, these results underscore the difficulty of the task, as alignment link presupposes accurate segmentation, which, as we saw, is not guaranteed due to the task ambiguities.

\subsection{Analysis of length differences}

Since interpreting typically produces shorter output than the input speech, we analyze this phenomenon from several perspectives: span length, relay (indirect) interpreting, and multi-track interpreting.

\begin{figure}[t]
    \centering
    \includegraphics[width=0.535\linewidth]{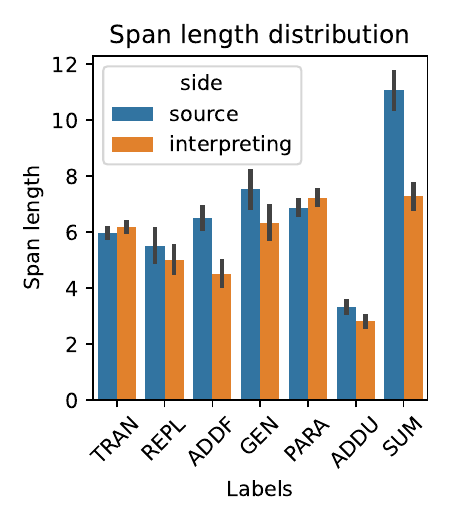}
    \includegraphics[width=0.445\linewidth]{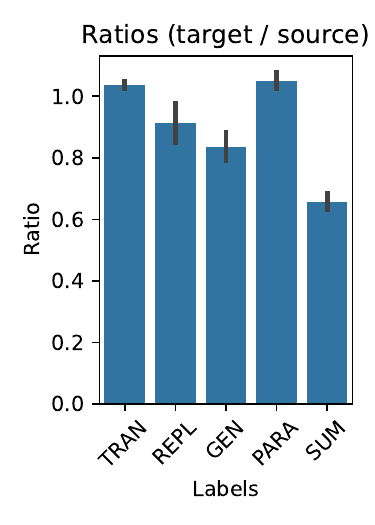}
    \caption{Left: Span length (in tokens) distribution per label for both source and target sides; Right: Weighted average of span length ratios (target / source) per label.}
    \label{fig:labels_statistics}
\end{figure}

\paragraph{Spans} \cref{fig:labels_statistics} (left) displays the distribution of span lengths (in tokens). The distribution seems to be uniform, except for \emph{uninformative additions}. Further inspection of \emph{additions} reveals that they are shorter because they contain only filler words, incomplete words or words such as ``very'', ``much'' etc. This figure also suggests that there is clear shortening happening in pairs of segments labeled \emph{summarization}. We thus plot the weighted average (with weights corresponding to the word counts in the source segment) of ratios of the target and source span length. We use a weighted average to make longer segments contribute more since the ratio in short segments can be caused only by the grammatical properties of language (e.g.\ articles in the English text that are not present in Czech).

\cref{fig:labels_statistics} (right) displays length ratios for each span label.\footnote{\change{We do not display ADDU and ADDF, as additions lack the counterpart for comparison.}} We see that the ratios for translation and paraphrase are very close to~1, as expected. Another observation is that length ratios for \emph{generalization} and \emph{summarization} are lower than one: 0.9 and 0.6 on average, respectively. This also aligns well with our intuition.

\paragraph{Relay interpreting}

Our corpus contains 27 direct interpretations and 12 indirect (relay) interpretations. On average, the ratio of source length to interpreting length, measured in characters,
is 77.5\% for direct interpreting and 97.43\% for relay interpreting. This suggests that relay interpreting may be somewhat easier than direct interpreting, as the first interpreter often already simplifies the content. Additionally, we observe a higher proportion of translations and fewer additions in relay interpreting. Further details are in \cref{app:direct_vs_indirect}.

\paragraph{Multi-track interpreting}

Another interesting feature of our interpreting dataset is the inclusion of multi-track interpreting, where the same speech is interpreted into the same language by two interpreters. We identified 7 such pairs and computed the average length ratio at both the character and token levels. On average, such pairs of interpretations differ by only 2\%, but the \emph{maximum} difference reaches 15\% for characters and 10\% for tokens. Detailed statistics are in \cref{app:multi_track}.

\subsection{Errors in interpreting \label{ssec:span-alignments}}

\begin{table}[t]
    \centering
    \footnotesize
    \setlength{\tabcolsep}{2.3pt}
    \begin{tabular}{l|ccccccc}
         & TRAN  & PARA & SUM & ADDF & GEN & ADDU & REPL \\\hline
source & 42.82  & 17.91 & 11.89 & 13.28 & 4.68 & 5.45 & 3.96 \\
target & 52.16  & 22.08 & 9.07 & 4.02 & 4.57 & 3.91 & 4.18 \\
    \end{tabular}
    \caption{The percentage of tokens with respective labels in the source and target side.}
    \label{tab:tokens_percentage}
\end{table}

We study the coverage of spans with respect to the distribution of labels to analyze potential errors and discrepancies. In \cref{tab:tokens_percentage}, we report the number of tokens belonging to each span label for both the source and the interpreting sides. The most frequent span label is \emph{translation}, which makes up for approximately half of all cases. The second is \emph{paraphrase}, accounting for one fifth. These results are in line with our intuition. We also observe that 13.3\% of tokens belong to spans where a factual omission is detected. Interestingly, there are also some factual additions in the target speech. We hypothesize it might happen when the interpreter misunderstands some part of the speech, but given the context, it is not suitable to label it as a \emph{replacement}. Some examples are discussed in \cref{sec:examples}.

\subsection{Examples \label{sec:examples}}

\begin{table*}[t]
    \centering
    \begin{tabular}{l|p{0.85\textwidth}}
        Label & Example (source speech $\to$ target speech) \\\hline
        TRAN & share the screen with my presentation $\to$ \it share the screen with my presentation\\
        PARA & No one can predict what will or won't happen $\to$ \it Because many things can happen \\
        SUM & And what can you do as an expectant mother? $\to$ \it As for mothers \\
        GEN & gynaecologist $\to$ \textit{doctor}; abuse $\to$ \it rude behavior \\
        REPL & 36.1 $\to$ 36,9; 12.4 $\to$ 12; in 2005 or after , not before 2005 $\to$ \it from 2005 to 2016 \\
        ADDF &towards this artificial intelligence which didn't $\to$ towards \textit{this camera} and the artificial intelligence didn't \\
        ADDU & \it For example; Next; Okay; can be also seen; And obviously \\
    \end{tabular}
    \caption{Example alignment links and their labels. For illustration purposes, all texts are translated into English even though they occurred in a different language in the dataset. Parts in \textit{italics} denote spans that were marked with the corresponding line label.}
    \label{tab:alignment_examples}
\end{table*}

\cref{tab:alignment_examples} presents some examples of annotations. We observe that there are some factual additions in the interpreting. This happens in cases when an interpreter is influenced by the preceding context and repeats information that conflicts with the original speech. For instance, in one talk, the speaker mentioned ``camera'' in combination with ``artificial intelligence''. This was later brought up by the interpreter even though it was not mentioned in the corresponding speech segment.

\section{Towards automatic alignment \label{sec:experiments}}
In this section, we showcase the use of \DatasetName{} as a useful resource to develop and evaluate alignment tools for interpreting. We describe a baseline system computing annotations at the word and span levels, then propose metrics to measure its accuracy and finally highlight its limitations.

\subsection{Methodology}

We implemented a simple system for automatic alignment similar to the proposal of \citet{zhao-etal-2024-naist}, which operates in three steps: (1) coarse alignment, (2) sub-segmentation to identify span-aligned pairs (with word alignment links within them), and (3) assigning labels to the span-level alignment links.

\paragraph{Coarse Alignment}

The first step is to obtain a high-precision coarse alignment at the span level. For this, we use BERTAlign \citep{liu2023bertalign}, a sentence alignment tool, configured with the following parameters: \texttt{max\_align 10}, \texttt{top\_k 10}, \texttt{window 10}, \texttt{skip 0.0} and \texttt{len\_penalty}.\footnote{Refer to the original work for the parameter description.} We emphasize that this process produces n-m sentence alignments, as interpreting naturally deviates from the traditional 1-1 sentence alignment that is majoritary observed in textual parallel corpora. High precision is prioritized at this stage to ensure the quality of subsequent sub-segmentations. We denote the resulting system for this first step as \emph{\SystemName{}}.

\paragraph{Sub-segmentation}

We compute sub-seg\-men\-ta\-tion and word alignments simultaneously. First, we identify all word alignment links using the \texttt{itermax} strategy from \citet{jalili-sabet-etal-2020-simalign}, configured with zero distortion, and the XLM-R model for computing contextual word embeddings \cite{conneau-etal-2020-unsupervised}. Next, we refine the spans by splitting them at points where two punctuations align in the source and target transcripts. This step generates additional span-level alignment links with shorter spans, resulting in system \emph{\SystemSubName{}}.

\paragraph{Labeling}

As \change{ previous steps may generate additions (n-0 or 0-m alignments) and translations (n-m alignments), we label additions as ADDU as it is the most frequent subcategory, and translations simply as TRAN.} \change{To also predict the other labels}, we implemented a very simple classification model in PyTorch\footnote{\url{https://pytorch.org/}} which takes the similarity score calculated by the multilingual sentence embedder LaBSE \citep{feng-etal-2022-language}, taking source and target span length as input features. It passes them through two hidden fully connected layers of size 100 and classifies the output into 5 categories, resulting in the system denoted \emph{\SystemLabName{}}. \change{Since we do not have training data yet, we use devset for training where we take 80\% of devset for actual training and 20\% as held-out data for the evaluation during the training.}

\subsection{Metrics \label{sec:metrics}}
The tasks considered in this work combine three difficulties: (a) to find the right spans, both in the source and the target; (b) to identify the correct alignment links between these spans, and with them the correct word alignments; (c) to label the links with their appropriate type. Our evaluation metrics take these three aspects into account. 

\paragraph{Segmentation} We evaluate the quality of span splits using accuracy, precision, recall, and $F_1$ of span boundaries, separately for the source and target texts. To also reward segmentation boundaries that are almost correct, we consider less severe metrics such as $P_k$ \cite{beeferman-lafferty-1999-statistical} and WindowDiff \citep{pevzner-hearst-2002-critique}. $P_k$ works by sliding a window of size $k$ over the text, comparing whether pairs of words at the boundaries of the window fall within the same span or not in both the source and target language. WindowDiff focuses on comparing the number of boundaries within a sliding window of size $k$. In practice, $k$ is set to the half of the average span size in the reference ($k=3$ in our case). Both metrics report a probability of error, with lower values corresponding to better segmentation. We use the NLTK implementation of these metrics \cite{nltk}.\footnote{\url{https://www.nltk.org}}

\paragraph{Span and word Alignment} We compute the proportion of exact matches for span alignment, which we call \emph{Exact match}. We distinguish between matching both span boundaries and labels or only span boundaries. As this metric is very strict, we also define an approximate span alignment error, which, similarly to the sentence alignment error \cite{veronis-langlais-2000-evaluation}, takes near-miss into account. This is computed as follows: for each pair of segments $(s,t)$ occurring in the reference or hypothesis alignment, we compute a list containing all possible word pairs $(u,v)$ with $u \in s$ and $v \in t$. Taking the union of such lists over the reference and hypothesis alignment yields two lists of word-level links, from which we compute Precision, Recall, and $F_1$. We refer to this metric as \emph{Relaxed match}. For the word alignment, we report Alignment Error Rate (AER) and $F_1$ macro-averaged over all recordings. These scores are computed with the implementation of \citet{azadi-etal-2023-pmi}.

\paragraph{Label match}
Given the difficulty of obtaining high segmentation quality and exact matches for alignment links, we only evaluate label correctness at the token level: Each token is labeled like the span it belongs to, and we then assess the proportion of correct link labels using accuracy and $F_1$.

\subsection{Baselines \label{ssec:baselines}}

\paragraph{Span alignment baseline} For the evaluation of segmentation, span alignment, and labeling, we compare \SystemName~to a random baseline, which randomly selects \change{the same number of boundaries in the source (resp.\ target) sides compared to the reference alignment,} and iterates through segments on both sides in parallel from left to right, randomly selecting a link label \change{from the shuffled pool of reference alignment links. This ensures that the number of labels of each type is the same as in the reference. Note that if the label is ADDU or ADDF, the span on only one side is labeled; otherwise the alignment link is created.}

\paragraph{Word alignment baseline} For word alignments, we use SimAlign \cite{jalili-sabet-etal-2020-simalign} as a baseline applied to the whole set of transcripts. We compute contextual embeddings using a sliding window of size 128 with stride 64. We discard links that connect words further than 50 tokens away, i.e.\ given source word $w_s$ and target word $w_t$ with their respective positions $p_{w_s}$ and $p_{w_t}$, we discard links if $|p_{w_s} - p_{w_t}| > 50$.

\subsection{Results \label{ssec:results}}

\begin{table*}[t]
    \centering
    \footnotesize
    \setlength{\tabcolsep}{3pt}
    \begin{tabular}{c|l|ccccc|ccc|cc|cc|cc|cc}
    && \multicolumn{5}{c|}{Segmentation} & \multicolumn{3}{c|}{Relaxed match} & \multicolumn{2}{c|}{Exact match} & \multicolumn{2}{c|}{Word align.} & \multicolumn{2}{c|}{Label match} & \multicolumn{2}{c}{\#span}\\\hline
& sys$_{annotator}$ & $P$$\uparrow$ & $R$$\uparrow$ & $F_1$$\uparrow$ & Df$\downarrow$ & $P_k$$\downarrow$ & $P$$\uparrow$ & $R$$\uparrow$ & $F_1$$\uparrow$ & w/$\uparrow$ & w/o$\uparrow$ & AER$\downarrow$ & $F_1$$\uparrow$ & acc$\uparrow$ & $F_1$$\uparrow$ & src & tgt \\\hline\hline
\parbox[t]{2mm}{\multirow{8}{*}{\rotatebox[origin=c]{90}{1 recording}}} & \BaselineName{}$_2$ & 15.12 & 14.44 & 14.77 & 0.50 & 0.47 & 0.08 & 0.06 & 0.07 & 0.00 & 0.00 & 0.74 & 0.25 & 55.35 & 56.65 & \multirow{2}{*}{145} & \multirow{2}{*}{123}\\
 & \BaselineName{}$_3$ & 21.32 & 15.80 & 18.15 & 0.54 & 0.49 & 0.05 & 0.04 & 0.04 & 0.00 & 0.00 & 0.70 & 0.28 & 36.00 & 29.48 & & \\ \cdashline{2-18}
 & \SystemName{}$_2$ & 97.37 & 41.11 & 57.81 & 0.23 & 0.23 & 0.43 & 0.99 & 0.60 & 10.60 & 14.57 & 0.30 & \bf 0.71 & \bf 76.20 & 67.73 & \multirow{2}{*}{59} & \multirow{2}{*}{53}\\
 & \SystemName{}$_3$ & \bf 98.25 & 32.18 & 48.48 & 0.33 & 0.30 & 0.39 & \bf 1.00 & 0.56 & 2.97 & 10.40 & 0.36 & 0.65 & 48.74 & 34.05 & & \\ \cdashline{2-18}
 & \SystemSubName{}$_2$ & 86.67 & 52.96 & \bf 65.75 & \bf 0.21 & \bf 0.20 & 0.52 & 0.82 & 0.63 & 15.89 & 18.54 & 0.34 & 0.66 & \bf 76.20 & 67.73 & \multirow{2}{*}{87} & \multirow{2}{*}{76}\\
 & \SystemSubName{}$_3$ & 85.45 & 40.52 & 54.97 & 0.32 & 0.28 & 0.45 & 0.80 & 0.58 & 4.46 & 11.39 & 0.40 & 0.61 & 48.74 & 34.05 & & \\ \cdashline{2-18}
 & \SystemLabName{}$_2$ & 86.67 & 52.96 & \bf 65.75 & \bf 0.21 & \bf 0.20 & 0.52 & 0.82 & 0.63 & 15.89 & 18.54 & 0.34 & 0.66 & 72.43 & \bf 70.36 & \multirow{2}{*}{87} & \multirow{2}{*}{76} \\
 & \SystemLabName{}$_3$ & 85.45 & 40.52 & 54.97 & 0.32 & 0.28 & 0.45 & 0.80 & 0.58 & 4.95 & 11.39 & 0.40 & 0.61 & 47.61 & 37.46 & & \\ \cdashline{2-18}
 & Annotator3$_2$ & 56.61 & \bf 72.96 & 63.75 & 0.30 & 0.25 & \bf 0.78 & 0.70 & \bf 0.74 & \bf 19.87 & \bf 30.46 & \bf 0.28 & \bf 0.71 & 57.60 & 65.39 & 184 & 159\\ \cdashline{2-18}
 & Annotator2$_3$ & 72.96 & 56.61 & 63.75 & 0.30 & 0.25 & 0.70 & 0.78 & \bf 0.74 & 14.85 & 24.26 & 0.36 & 0.66 & 57.60 & 49.82 & 145 & 123\\ \hline
\parbox[t]{2mm}{\multirow{4}{*}{\rotatebox[origin=c]{90}{devset}}} & \BaselineName{} & 17.16 & 16.07 & 16.60 & 0.47 & 0.43 & 0.04 & 0.03 & 0.03 & 0.14 & 0.18 & 0.70 & 0.30 & 36.18 & 37.93 & 195 & 176\\
 & \SystemName{} & \bf 95.59 & 35.33 & 51.59 & 0.25 & 0.23 & 0.38 & \bf 0.97 & 0.54 & 6.83 & 11.98 & \bf 0.32 & \bf 0.69 & \bf 58.52 & 44.98 & 72 & 64\\
 & \SystemSubName{} & 79.45 & \bf 50.04 & \bf 61.40 & \bf 0.24 & \bf 0.21 & \bf 0.51 & 0.71 & \bf 0.60 & \bf 9.70 & \bf 16.44 & 0.38 & 0.63 & 58.48 & 44.98 & 125 & 107\\
 & \SystemLabName{} & 79.45 & \bf 50.04 & \bf 61.40 & \bf 0.24 & \bf 0.21 & \bf 0.51 & 0.71 & \bf 0.60 & 9.61 & \bf 16.44 & 0.38 & 0.63 & 52.25 & \bf 47.68 & 125 & 107\\ \hline
\parbox[t]{2mm}{\multirow{4}{*}{\rotatebox[origin=c]{90}{testset}}} & \BaselineName{} & 19.34 & 17.79 & 18.53 & 0.51 & 0.45 & 0.05 & 0.03 & 0.04 & 0.14 & 0.24 & 0.75 & 0.27 & 26.86 & 27.70 & 213 & 185\\
 & \SystemName{} & \bf 95.05 & 28.26 & 43.56 & 0.31 & 0.30 & 0.28 & \bf 0.95 & 0.44 & 4.21 & 10.39 & \bf 0.37 & \bf 0.65 & \bf 41.24 & 25.92 & 62 & 59\\
 & \SystemSubName{} & 82.52 & \bf 43.43 & \bf 56.91 & \bf 0.28 & \bf 0.25 & \bf 0.44 & 0.74 & 0.55 & 6.41 & \bf 13.80 & 0.42 & 0.59 & 41.22 & 25.99 & 110 & 104\\
 & \SystemLabName{} & 82.52 & \bf 43.43 & \bf 56.91 & \bf 0.28 & \bf 0.25 & \bf 0.44 & 0.74 & \bf 0.55 & \bf 6.55 & \bf 13.80 & 0.42 & 0.59 & 38.16 & \bf 31.91 & 110 & 104\\
\end{tabular}
    \caption{The evaluation of our system is detailed as follows: w/ and w/o in the Exact Match evaluation represent results with and without labels, respectively. \#span represents the average span count for each split. BA refers to the system after applying BERTAlign (the first step), +sub indicates the BA system extended with follow-up sub-segmentation (the second step), and +lab represents the system further enhanced by labeling (the third step). \change{For 1 recording, the subscript indicates the ID of the annotator whose annotation is used for evaluating the alignment.}}
    \label{tab:results}
\end{table*}

We evaluate the random baseline and our systems on three dataset splits: (1) one recording for which a double annotation is available; (2) development set and (3) test set. \change{The alignments for (1) are evaluated separately for each annotator, with the annotator ID is indicated as a subscript.} The results are in \cref{tab:results} and further detailed below:

\paragraph{Segmentation}

The first block of \cref{tab:results} presents the evaluation of segmentation quality. As intended in the first step, \SystemName{} demonstrates a very high precision. While sub-segmentation slightly reduces precision, it improves the overall $F_1$ score. Notably, \SystemSubName{} even surpasses annotator 2 in inter-annotator comparisons, as reflected in both the $F_1$ score and in metrics such as Window Diff and $P_k$.

\paragraph{Span and Word Alignment}

The second and third blocks of \cref{tab:results} report the quality of relaxed and exact matches for predicted span alignment links, respectively. For relaxed matches, \SystemSubName{} performs slightly below the level of inter-annotator agreement. In the case of exact matches (third block), performance varies depending on the comparison with Annotator 1 or Annotator 2. This difference can be attributed to the number of alignment links: Annotator 1 (143 links) aligns more closely with \SystemName{} (90 links) compared to Annotator 2. The fourth block of \cref{tab:results} evaluates the quality of word alignment links, showing that traditional word alignment tools designed for MT struggle due to the longer context in interpreting. Even with a moving window that discards distant links, the baseline approach performs significantly worse than our method.

\paragraph{Label Match}

The final block of \cref{tab:results} reports the quality of per-token annotation labels. While label classification improves upon the default label prediction, the improvement is modest. This suggests that segmenting solely based on punctuations inserted in the transcription phase is insufficient for interpreting, highlighting the need for a more fine-grained solution. We leave this for future work.

\section{Related work \label{sec:related}}
\paragraph{Sentence alignment}
Sentence-aligned corpora are key to modern MT and have been studied since statistical MT emerged \cite{tiedemann-2011-bitext}. Their mostly monotonic, 1-to-1 nature makes alignment computationally efficient, enabling large parallel data repositories like Opus \cite{tiedemann-2012-parallel}.\footnote{\url{https://opus.nlpl.eu}}

\paragraph{Word alignment} Word alignment annotation has been widely studied, starting with the Bible \cite{melamed1998annotation} and the Canadian Hansards proceedings \cite{och2000improved}, then expanding to more languages, mostly paired with English: Romanian, Hindi, Inuktitut \cite{martin-etal-2003-aligning}, Spanish \cite{lambert2005guidelines}, Czech \cite{bojar-prokopova-2006-czech,kruijff-korbayova-etal-2006-annotation}, and Portuguese \cite{graca-etal-2008-building}, etc. These alignments are typically ``flat'', linking words directly. More complex alignments, mapping nodes in parallel parse trees, exist for Japanese, Chinese \cite{uchimoto-etal-2004-multilingual}, German \cite{volk-etal-2006-xml}, Danish \cite{buch-kromann-2007-computing}, and Chinese and Arabic (Gale project) \cite{li2010parallel}. The Czech-English parallel dependency treebank \cite{hajic-etal-2012-announcing} also provides large-scale automatic annotations. Such annotations capture not only word correspondences but also syntax-level equivalences. Hierarchical span alignments have been manually annotated for French using an iterative divisive procedure \cite{xu-yvon-2016-novel}. These works inspired our annotation guidelines (\cref{app:guidelines}). While most word alignments focus on written texts, speech data remains underexplored, except for broadcast news transcripts in the Gale project \cite{li2010parallel}.


\paragraph{Interpreting Datasets}

Several simultaneous interpreting corpora exist, including EPIC \cite{sandrelli-bendazzoli-2006-tagging}, EPIC-Ghent \cite{defrancq2015corpus}, and EPTIC \cite{bernardini2016epic}, which are small collections of transcribed European Parliament interpretations for analysis. Additional corpora have been published by \citet{temnikova-etal-2017-interpreting, pan-2019-chinese}. The ESIC corpus \cite{machacek21_interspeech} covers multiple languages and includes transcripts, translations, and simultaneous interpreting transcripts. Other resources, mainly for consecutive interpreting, are documented by \citet{lazaro-gutierrez-2023-natural}. However, none of these corpora provide alignments between speeches.

\paragraph{Alignment annotation in interpreting} \citet{doi-etal-2021-large} present a large-scale (around 300 hours) English-Japanese simultaneous interpretation corpus along with the results of its analysis. Part of the dataset is manually annotated (14 TED talks) with categories such as additions, pragmatically uninformative omissions, and factual omissions. They further evaluate the dataset based on latency, quality, and word order. Building on this corpus, \citet{zhao-etal-2024-naist} provide an automatically aligned parallel English-Japanese interpretation dataset. Their approach, similar to ours, involves two steps: coarse alignment followed by fine-grained alignment. Their error analysis addresses unintentional omissions (corresponding to our ``additions'' in source speech), intentional omissions (summarization), and mistranslations (replacements).

\section{Conclusion \label{sec:conclusion}}


In this paper, we have detailed our efforts to collect, prepare and annotate a corpus of simultaneous interpretings, performed by student interpreters in mock conferences. We discussed the guidelines used at each annotation step and reported the results of the first analysis of the resulting corpus. They illustrate how interpreting activities could be studied and monitored with corpus-based techniques; they also highlight the need to develop dedicated tools for their annotation. The resulting corpus and tools will be released to the community. In a final step, we used this new resource to evaluate automatic alignment tools for interpreting corpus: as it seems, this new task, which combines the difficulties of multiple existing annotation processes, poses challenges for our existing alignment tools.

\paragraph{Future work}
We plan to deepen our preliminary observations at several levels: to better correlate the main speaker's oral production with labels on the interpreting side; to also study how interpreting strategies vary depending on the source and target languages.
A lot finally remains to be done to improve our automatic processing tools which do not rely on punctuation as it is a very unreliable alignment indicator in interpreting.

\section*{Limitations}
We acknowledge that the current dataset is only limited in size and linguistic diversity, which is hardly compensated by the richness of available annotations. We are continuously working on extending this dataset, with the hope of accumulating a sufficiently large set of annotated speeches that could also be used for training (or fine-tuning) a supervised machine learning system and improving the automatic span-level annotations. Regarding the alignment tool, an obvious limitation is the lack of connection with the original speech, which needs to be transcribed by an external tool, then revised, before the alignment takes place. As a first step towards a tighter integration, we could work on providing the annotators with an integrated player, providing them with a way to listen to the original audio tracks and even correct the corresponding transcripts. \change{We additionally emphasize that the Random baseline does not uses Reformulation or Replacement labels and our approach is suboptimal the second phrase where we sub-segment on the punctuation match.}

\paragraph{Lower Inter-Annotator Agreement}  \change{We consider an annotation ``correct'' when annotators agree. In an ideal scenario, annotators would discuss and align their approaches during annotation. However, we found this setup both time-consuming and impractical. Additionally, defining the distinction between paraphrase and non-paraphrase is inherently challenging. While introducing minimal blocks corresponding to syntactic units might be a potential direction, \citet{leech2000grammars} has shown that syntax is not a good indicator of units in speech. Currently, we provided feedback on how well annotators adhered to guidelines after they annotated a part of the data. Despite these efforts, some divergence remains, reflecting the complexity of the task.}

\section*{Ethics Statement}
All data contained in the \DatasetName{} dataset are fully anonymized, e.i. they do not contain any personal information (names) about the speakers. We collected consents from speakers to publish recordings containing their voice and the transcripts of their speech. The participants were informed that their recordings will be used for research purposes.

\section*{Acknowledgements}

\change{The work has been partially supported by the grant 272323 of the Grant Agency of Charles University, SVV project number 260 821, the grant CZ.02.01.01/00/23\_020/0008518 (``Jazykověda, umělá inteligence a jazykové a řečové technologie: od výzkumu k aplikacím''), and by the funds provided by the French-Czech Barrande Fellowship Programme. Part of this work has been done while the first author was visiting the Institute for Intelligent Systems and Robotics (ISIR) at Sorbonne Université in Paris, France. We would like to express our sincere gratitude to prof. PhDr. Ivana Čeňková, CSc., Mgr. Kateřina Ešnerová, and all the students who generously gave their consent to have their interpreting data used in this study.}

\bibliography{anthology,custom}

\appendix

\section{Details about the \DatasetName{} dataset}

\begin{table*}[t]
    \centering
    \footnotesize
    \begin{tabular}{l|c|c|c|c|cc|ccc}
 & \multicolumn{2}{c|}{Language} &  \multicolumn{2}{c|}{Interpreting}  &  \multicolumn{2}{c}{Annotator} & \multicolumn{3}{|c}{Recording}  \\\hline
split & src & trg & relay & interpreter id & consent & annotator id & src id & trg id & duration \\\hline\hline
\multirow[c]{10}{*}{dev} & \multirow[c]{10}{*}{cs} & \multirow[c]{2}{*}{de} & no & 8& 1 & es & 9 & 10 & 00:11:21 \\
\cline{4-10} \cline{5-10} \cline{6-10} \cline{7-10} \cline{8-10}
 &  &  & yes & 8& 1 & de & 11 & 12 & 00:09:47 \\
\cline{3-10} \cline{4-10} \cline{5-10} \cline{6-10} \cline{7-10} \cline{8-10}
 &  & \multirow[c]{6}{*}{en} & \multirow[c]{4}{*}{no} & \multirow[c]{3}{*}{1 } & \multirow[c]{3}{*}{3} & \multirow[c]{3}{*}{en 1} & 5 & 6 & 00:12:31 \\
\cline{8-10}
 &  &  &  &  &  &  & 7 & 8 & 00:09:04 \\
\cline{8-10}
 &  &  &  &  &  &  & 16 & 17 & 00:13:52 \\
\cline{5-10} \cline{6-10} \cline{7-10} \cline{8-10}
 &  &  &  & 6 & 3 & en 2 & 9 & 13 & 00:11:09 \\
\cline{4-10} \cline{5-10} \cline{6-10} \cline{7-10} \cline{8-10}
 &  &  & \multirow[c]{2}{*}{yes} & 11 & 3 & en 1 & 3 & 15 & 00:10:15 \\
\cline{5-10} \cline{6-10} \cline{7-10} \cline{8-10}
 &  &  &  & 6 & 3 & en 1 & 3 & 4 & 00:10:05 \\
\cline{3-10} \cline{4-10} \cline{5-10} \cline{6-10} \cline{7-10} \cline{8-10}
 &  & es & no & 4  & 3 & es & 9 & 14 & 00:11:20 \\
\cline{3-10} \cline{4-10} \cline{5-10} \cline{6-10} \cline{7-10} \cline{8-10}
 &  & fr & yes & 3& 1 & fr & 1 & 2 & 00:20:07 \\\hline\hline
\multirow[c]{29}{*}{test} & \multirow[c]{15}{*}{cs} & \multirow[c]{3}{*}{de} & no & 5& 3 & de & 9 & 46 & 00:11:13 \\
\cline{4-10} \cline{5-10} \cline{6-10} \cline{7-10} \cline{8-10}
 &  &  & \multirow[c]{2}{*}{yes} & 12& 1 & de & 57 & 58 & 00:09:27 \\
\cline{5-10} \cline{6-10} \cline{7-10} \cline{8-10}
 &  &  &  & 5& 3 & de & 11 & 39 & 00:09:47 \\
\cline{3-10} \cline{4-10} \cline{5-10} \cline{6-10} \cline{7-10} \cline{8-10}
 &  & \multirow[c]{6}{*}{en} & \multirow[c]{3}{*}{no} & 11 & 3 & en 2 & 9 & 45 & 00:10:59 \\
\cline{5-10} \cline{6-10} \cline{7-10} \cline{8-10}
 &  &  &  & \multirow[c]{2}{*}{2 } & \multirow[c]{2}{*}{3} & \multirow[c]{2}{*}{en 1} & 7 & 18 & 00:09:04 \\
\cline{8-10}
 &  &  &  &  &  &  & 30 & 31 & 00:11:30 \\
\cline{4-10} \cline{5-10} \cline{6-10} \cline{7-10} \cline{8-10}
 &  &  & \multirow[c]{3}{*}{yes} & 6 & 3 & en 2 & 11 & 40 & 00:09:47 \\
\cline{5-10} \cline{6-10} \cline{7-10} \cline{8-10}
 &  &  &  & 7 & 3 & en 2 & 11 & 41 & 00:09:47 \\
\cline{5-10} \cline{6-10} \cline{7-10} \cline{8-10}
 &  &  &  & 9 & 3 & en 1 & 48 & 49 & 00:09:39 \\
\cline{3-10} \cline{4-10} \cline{5-10} \cline{6-10} \cline{7-10} \cline{8-10}
 &  & \multirow[c]{3}{*}{es} & \multirow[c]{3}{*}{yes} & 13  & 3 & es & 48 & 55 & 00:09:31 \\
\cline{5-10} \cline{6-10} \cline{7-10} \cline{8-10}
 &  &  &  & \multirow[c]{2}{*}{4 } & \multirow[c]{2}{*}{3} & \multirow[c]{2}{*}{es} & 3 & 51 & 00:10:20 \\
\cline{8-10}
 &  &  &  &  &  &  & 52 & 53 & 00:11:31 \\
\cline{3-10} \cline{4-10} \cline{5-10} \cline{6-10} \cline{7-10} \cline{8-10}
 &  & \multirow[c]{3}{*}{fr} & \multirow[c]{3}{*}{no} & \multirow[c]{3}{*}{3 } & \multirow[c]{3}{*}{1} & \multirow[c]{3}{*}{fr} & 5 & 27 & 00:12:31 \\
\cline{8-10}
 &  &  &  &  &  &  & 7 & 19 & 00:09:04 \\
\cline{8-10}
 &  &  &  &  &  &  & 34 & 35 & 00:07:54 \\
\cline{2-10} \cline{3-10} \cline{4-10} \cline{5-10} \cline{6-10} \cline{7-10} \cline{8-10}
 & \multirow[c]{2}{*}{de} & \multirow[c]{2}{*}{cs} & \multirow[c]{2}{*}{no} & 12  & 1 & de & 47 & 48 & 00:09:37 \\
\cline{5-10} \cline{6-10} \cline{7-10} \cline{8-10}
 &  &  &  & 8  & 1 & de & 54 & 52 & 00:11:37 \\
\cline{2-10} \cline{3-10} \cline{4-10} \cline{5-10} \cline{6-10} \cline{7-10} \cline{8-10}
 & \multirow[c]{5}{*}{en} & \multirow[c]{5}{*}{cs} & \multirow[c]{5}{*}{no} & 10  & 3 & en 2 & 42 & 44 & 00:09:25 \\
\cline{5-10} \cline{6-10} \cline{7-10} \cline{8-10}
 &  &  &  & \multirow[c]{3}{*}{2 } & \multirow[c]{3}{*}{3} & \multirow[c]{3}{*}{en 2} & 22 & 23 & 00:09:42 \\
\cline{8-10}
 &  &  &  &  &  &  & 26 & 1 & 00:20:07 \\
\cline{8-10}
 &  &  &  &  &  &  & 36 & 37 & 00:13:48 \\
\cline{5-10} \cline{6-10} \cline{7-10} \cline{8-10}
 &  &  &  & 9  & 3 & en 2 & 42 & 43 & 00:09:25 \\
\cline{2-10} \cline{3-10} \cline{4-10} \cline{5-10} \cline{6-10} \cline{7-10} \cline{8-10}
 & \multirow[c]{2}{*}{es} & \multirow[c]{2}{*}{cs} & \multirow[c]{2}{*}{no} & 13  & 3 & es & 56 & 57 & 00:09:32 \\
\cline{5-10} \cline{6-10} \cline{7-10} \cline{8-10}
 &  &  &  & 4  & 3 & es & 38 & 11 & 00:09:47 \\
\cline{2-10} \cline{3-10} \cline{4-10} \cline{5-10} \cline{6-10} \cline{7-10} \cline{8-10}
 & \multirow[c]{5}{*}{fr} & \multirow[c]{5}{*}{cs} & \multirow[c]{5}{*}{no} & 14  & 3 & fr & 50 & 3 & 00:09:53 \\
\cline{5-10} \cline{6-10} \cline{7-10} \cline{8-10}
 &  &  &  & \multirow[c]{4}{*}{3 } & \multirow[c]{4}{*}{1} & \multirow[c]{4}{*}{fr} & 20 & 21 & 00:09:10 \\
\cline{8-10}
 &  &  &  &  &  &  & 24 & 25 & 00:07:59 \\
\cline{8-10}
 &  &  &  &  &  &  & 28 & 29 & 00:08:32 \\
\cline{8-10}
 &  &  &  &  &  &  & 32 & 33 & 00:10:38 \\
\cline{1-10} \cline{2-10} \cline{3-10} \cline{4-10} \cline{5-10} \cline{6-10} \cline{7-10} \cline{8-10}
\end{tabular}
    \caption{Detailed statistics of \DatasetName{}. Consent values 1 and 3 denote consents to publish only transcripts or both the transcripts and audio, respectively.}
    \label{tab:detailed_statistics_dataset}
\end{table*}

\subsection{Statistics}
\label{app:detailed_statistics}
Detailed statistics of our \DatasetName{} dataset are presented in \cref{tab:detailed_statistics_dataset}.

\subsection{Topics}
\label{app:domains}

\begin{table*}[t]
    \centering
    \setlength{\tabcolsep}{3.5pt}
    \footnotesize
    \begin{tabular}{c|l}
    doc id & topic \\\hline
1 & Prevention of Traumatic Birth Experiences \\
3 & From Maison des Cultures du Monde: The Scope of Work of This Institution \\
5 & What Are the Benefits of Hypnobirthing \\
7 & The Brain Is Not a Computer \\
9 & A Cultural Anthropologist and Ethnologist Based at the University of Plzeň \\
11 & From Yucatan University: Mayan Script and Its Decipherment \\
16 & Harnessing Modern Technologies to Achieve Sustainable Development Goals \\
20 & Utilization of AI in the Military Field \\
22 & Scottish Inspiration for Prague \\
24 & Shift Moonwalkers - The Future of Walking? \\
26 & Prevention of Traumatic Birth Experiences \\
28 & School Transport: Pedibus \\
30 & Traffic Snake Game: Achieving Sustainable Mobility Through a Game \\
32 & Que Choisir: Activities and Mission of This Association \\
34 & Consumer Rights in the Past and Present and the Goals and Role of the dTest Organization \\
36 & Regulating Ads in the Digital Age: An Impossible Task \\
38 & From Yucatan University: Mayan Script and Its Decipherment \\
42 & On Freelance Business Development: Benefits of Cultural Diversity in the Workplace \\
47 & Team Leader of Charta der Vielfalt (Diversity Charter): Goals of the Charter and Activities of the Association \\
50 & From Maison des Cultures du Monde: The Scope of Work of This Institution \\
52 & Antigypsyism – History of Antigypsyism in Europe, Personal Experiences, Possible Solutions \\
56 & From the Spanish Organization Unión Romaní: Antigypsyism and the Paradox of Tolerance During the Pandemic \\
    \end{tabular}
    \caption{The topics of the speeches are listed alongside their document IDs in the first column. These IDs correspond to those in \cref{tab:detailed_statistics_dataset}.}
    \label{tab:domains}
\end{table*}

The topics for each speech of our \DatasetName{} dataset are presented in \cref{tab:domains}.

\subsection{Direct vs. relay interpreting}
\label{app:direct_vs_indirect}

\begin{figure}
    \centering
    \includegraphics[width=\linewidth]{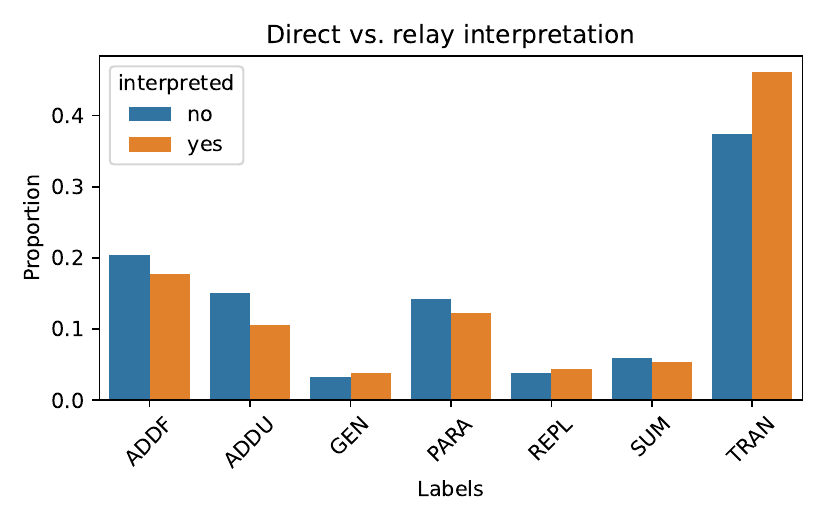}
    \caption{Relative proportion of each span label within each category (source interpreted: yes and no).}
    \label{fig:direct_relay_interpreting}
\end{figure}

\cref{fig:direct_relay_interpreting} presents the difference between direct and relay interpreting in terms of distribution of labels. we observe a higher proportion of translations and fewer additions in relay interpreting.

\subsection{Multi-track interpreting}
\label{app:multi_track}
\begin{table}[t]
    \centering
    \footnotesize
    \begin{tabular}{cc|cc}
    \multicolumn{2}{c|}{Document id} & \multicolumn{2}{c}{Ratio} \\\hline
    1. doc id & 2. doc id & character & token \\\hline\hline
    18 & 8 & 0.96 & 0.96 \\
    39 & 12 & 0.86 & 0.90 \\
    40 & 41 & 0.94 & 0.95 \\
    43 & 44 & 0.97 & 0.93 \\
    13 & 45 & 1.04 & 1.01 \\
    10 & 46 & 1.16 & 1.11 \\
    15 & 4 & 0.95 & 0.96 \\
    \end{tabular}
    \caption{Character and token ratios for multi-track interpreting. The first two columns denote ids of documents that are interpretations of the same speech. More details about the documents are in \cref{tab:detailed_statistics_dataset} and \cref{tab:domains}.}
    \label{tab:multi_track_ratios}
\end{table}

\cref{tab:multi_track_ratios} presents the length ratios calculated on characters and tokens for the pairs of interpretations that share the same speech.

\subsection{Annotator Disagreement Example}
\label{app:annotator_disagreement}

\begin{figure}
    \centering
    \includegraphics[width=\linewidth]{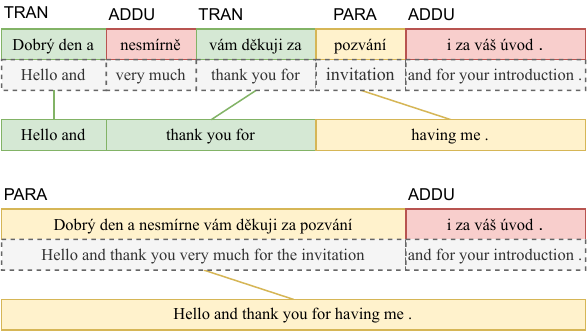}
    \caption{Two alignment annotations (by two different annotators) of the same sentence from the speech and its interpreting.}
    \label{fig:annotator_disagreement}
\end{figure}

\cref{fig:annotator_disagreement} illustrates the difference in annotation granularity that we discuss in \cref{sec:analysis}. The first row in \cref{fig:annotator_disagreement} is annotated by Annotator 3 and the second row by Annotator 2. We can see that Annotator 3 makes segment splits more often and produces a more fine-grained annotation, whereas Annotator 2 prefers longer segments.

\section{Annotation tools and \ToolName{}}
\label{app:annotation_tools}

\begin{table}[t]
    \centering
    \setlength{\tabcolsep}{3.5pt}
    \footnotesize
    \begin{tabular}{l|ccccc}
         Alignment annotation tool & \rotatebox[origin=c]{90}{Word-level} & \rotatebox[origin=c]{90}{Phrase-level} & \rotatebox[origin=c]{90}{Multilevel} & \rotatebox[origin=c]{90}{Long texts} & \rotatebox[origin=c]{90}{Modern} \\\hline\hline
         I*Link \cite{ahrenberg-etal-2003-interactive} & $\checkmark$ &&&&\\
         YAWAT \cite{germann-2008-yawat} & $\checkmark$ & $\checkmark$ &&& $\checkmark$\\
         Swift Aligner \cite{gilmanov-etal-2014-swift} & $\checkmark$ &&&&\\
         CLUE-Aligner \cite{barreiro2016clue} & $\checkmark$ & $\checkmark$ & $\checkmark$ && $\checkmark$\\
         MASSAlign \cite{paetzold-etal-2017-massalign} & $\checkmark$ & $\checkmark$ &&$\checkmark$&\\
         Line-a-line \cite{skeppstedt-etal-2020-line} & $\checkmark$ & $\checkmark$ &&& $\checkmark$ \\
         ManAlign \cite{steingrimsson-etal-2021-combalign} & $\checkmark$ &&&&\\
         Ugarit \cite{yousef-2022-ugarit} & $\checkmark$ & $\checkmark$ && $\checkmark$ & $\checkmark$ \\\hline
         InterAlign (ours) & $\checkmark$ & $\checkmark$ &$\checkmark$&$\checkmark$& $\checkmark$ \\
    \end{tabular}
    \caption{Existing alignment annotation tools and their main features.}
    \label{tab:tools}
\end{table}

\paragraph{Existing tools} We considered several existing tools: I*Link provides word-level alignment, compiling reports and statistics, and automatic proposals for token alignments \cite{ahrenberg-etal-2003-interactive}; YAWAT is a web-based tool for word and phrase-level alignments of parallel texts that are pre-segmented to sentences \cite{germann-2008-yawat}; Swift Aligner supports word-level alignment with additional capabilities for annotating dependency syntax and part-of-speech \cite{gilmanov-etal-2014-swift}; CLUE-Aligner is a web alignment tool designed for annotation of word or phrasal units in parallel sentences. \cite{barreiro2016clue}; MASSAlign is Python library for the alignment and annotation of monolingual comparable documents at word and sentence levels \cite{paetzold-etal-2017-massalign}; Line-a-line is a web-based tool for manual annotation of word-alignments in sentence-aligned parallel corpora \cite{skeppstedt-etal-2020-line}; AlignMan is a tool for manual word alignment of parallel sentences \cite{steingrimsson-etal-2021-combalign}; Ugarit is a public web-based tool for manual annotation of parallel texts for generating word- and phrase-level translation alignment, supporting the alignment between three parallel texts. A compact overview of all these tools is in \cref{tab:tools}.

\paragraph{Our requirements} Alignment of interpretings, however, differs from that of text translations, which is usually performed in two stages: first at the sentence, then at the word level. This is because interpretings do not include unambiguous sentence boundaries in their transcripts. Interpreters often also omit, or rephrase long spans, trying to jointly accommodate time and content-preservation constraints, making the resulting transcripts difficult to word-align.

Since we cannot rely on any prior sentence segmentation or sentence alignment between the source and interpreting, a strong requirement for us was to support the annotation of long spans comprising dozens of tokens. This narrowed our list of options down to practically one tool: Ugarit \cite{yousef-2022-ugarit}. Upon testing, we observed that it could not be used to perform both lexical and phrasal alignments at the same time.

\begin{figure*}[t]
    \centering
    \includegraphics[width=\textwidth]{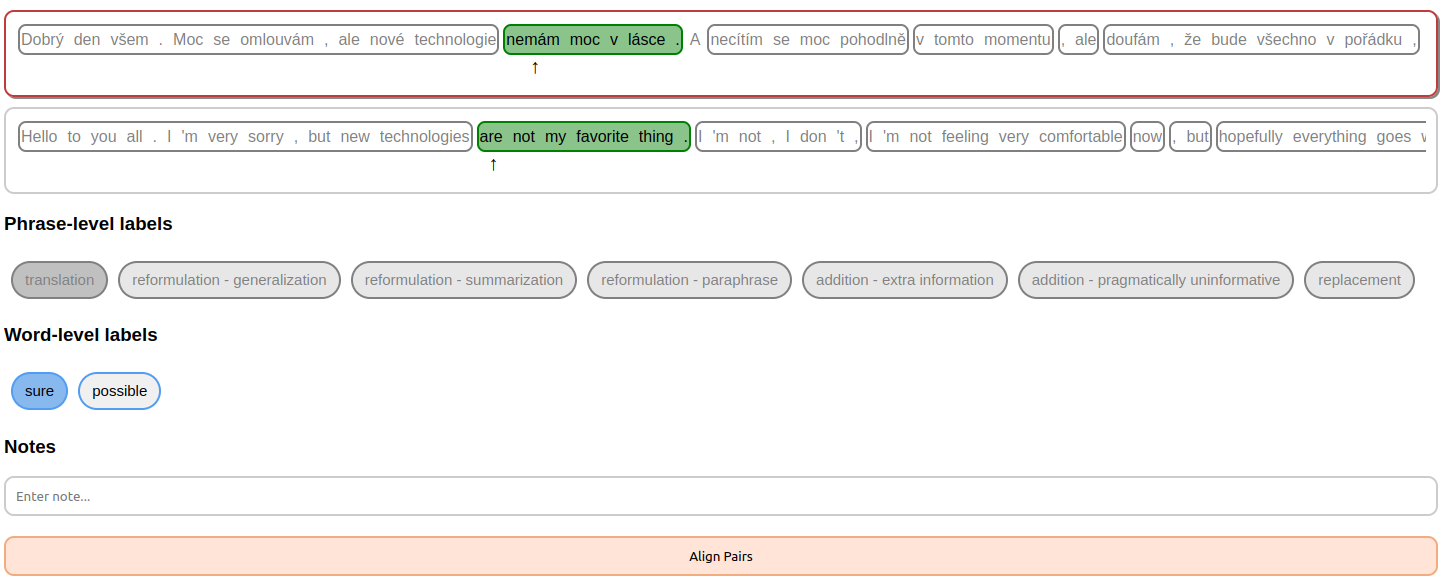}
    \caption{A screenshot of \ToolName{} for aligning transcripts of speech and their interpreting.}
    \label{fig:annotation-tool}
\end{figure*}

\paragraph{InterAlign} We, therefore, implemented a new annotation tool, \ToolName, that is primarily designed to be used for aligning transcripts of speech and their interpreting but can be used in any situation when no sentence segmentation and alignment is provided. It supports annotations at both the word- and span-level, can handle long texts, and enables the user to define its own span labels. The tool is implemented in React,\footnote{\url{https://react.dev/}} a modern web-based framework; it combines many individual features from previous annotation tools.

\begin{figure}[t]
    \centering
    \includegraphics[width=0.45\textwidth]{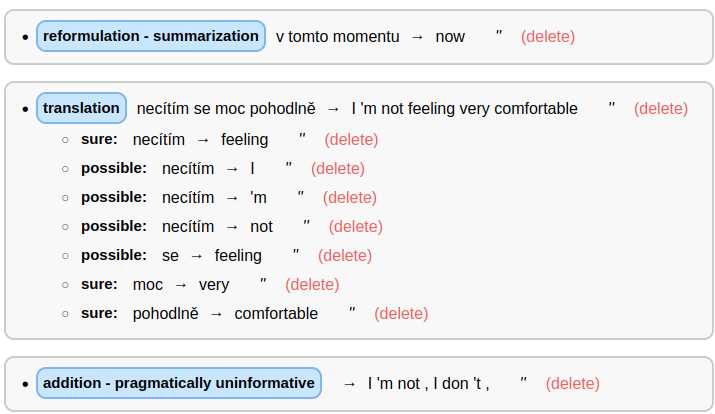}
    \caption{List of chunks and word alignment links displayed below the alignment window.}
    \label{fig:alignment-list}
\end{figure}

A screenshot of this tool is in \cref{fig:annotation-tool}. The transcripts are displayed horizontally in two scrollable elements, enabling the alignment of long chunks. Annotation links can be created either by both mouse or keyboard actions. After creating an alignment, the link is added to the list and displayed under the annotation interface. A screenshot of the link list is in \cref{fig:alignment-list}. 

\section{Transcript revision guidelines}
\label{app:transcript_guidelines}
\begin{enumerate}
    \item Please correct the transcripts to match what is said in the recordings.
    \begin{itemize}
        \item Do not correct grammar – if the speaker makes grammatical or any other language mistakes (stutters, repeats himself, uses the wrong form of a word or a whole word), the transcript should capture the exact notation of what is said.
\item For example, you can edit the stutter in the word international as: ``inter- international'' (with space between words).
\item Please record hesitations, interjections, etc. if they are obvious or inaudible. Please mark hesitations with @.
\item Please do not mark smacking and swallowing.
\item Please indicate a longer time delay in the speech with three dots.
    \end{itemize}
    \item You can change the segmentation to sentences.
    \begin{itemize}
        \item 
Transcripts already contain sentences. It is possible that a different sentence division is suitable, but you are welcome to create your own sentence division (but this is not required).
\item Please edit the sentences so that each one is on a separate line.
    \end{itemize}
    \item Label proper names.
    \begin{itemize}
        \item Recordings can contain the names of cities, organizations - it is important to mark these proper names with the [NAME] tag, for example, the sentence on the left will be the sentence on the right after the arrow: Václav was then in the Czech Republic. $\to$ [NAME](Václav) was then in [NAME](Czech Republic).
    \end{itemize}
\end{enumerate}

\section{Annotation Guidelines}
\label{app:guidelines}

\subsection{Phrase-level alignments}
\paragraph{Segmentation}
\begin{itemize}
    \item Divide the speech and its interpretation into segments that correspond to each other and label them with the following labels.
    \item Each segment's length should be maximal, meaning adding one more word to either side would change the label of the segment.
    \item Each word is assigned to exactly one segment.
\end{itemize}
\paragraph{Labeling Criteria}
Labels are assigned to the interpretation when you compare it to the source speech. For instance, ``summarization'' means that some part of the interpreting (the second transcript) is summarized given the original speech (the first transcript). Similarly, ``addition''
means that some information is added in the interpretation. More precisely, the labels are:
\begin{itemize}
    \item Translation: Direct translation that holds outside of any additional context.
    \item Reformulation:
    \begin{itemize}
        \item ``Paraphrase'': Equivalent meaning in the context, but not a direct translation.
        \item ``Summarization'': Equivalent meaning but the interpretation is expressed in less words, summarized.
        \item ``Generalization'': The meaning is as close as possible, but one side of the aligned pair is less specific. For instance, instead of saying “cats and dogs”, it is said ``pets''. Or instead of a particular name of a village, there is only ``some village'' mentioned.
    \end{itemize}
\item Addition: Used only on one of the sides, to indicate that this span brings additional content not present in the other language. Please distinguish the following sub-classes of ``addition'':
\begin{itemize}
    \item Extra information: the interpreting adds some new information, the meaning of the text is changed;
    \item Pragmatically uninformative: the interpreting does not change the meaning, the span repeats something that has already been said or is not related to the topic.
\end{itemize}
\item ``Replacement'': Obvious error, misunderstanding a number, place, name, etc. (e.g. instead of saying 17, it is said 70. In English it is very similar and it can be clear from the context that 70 is a replacement of 17)
\end{itemize}

\paragraph{Notes}
Make notes about any hesitations or uncertainties you may have during the annotation process.

\subsection{Priorities of Phrase-level Labels}
When considering which label to use for an aligned phrase pair, prefer segmentation and labels in this order:

\begin{enumerate}
    \item ``Translation'' (Alignment): If a word in the source span directly corresponds to a translation in the target span out of any additional context, mark it as a translation alignment. Ensure accuracy and precision in aligning words with their translations.
    \item ``Reformulation'':
    Identify phrases in the source span that convey the same meaning as phrases in the target span but are not direct translations. Use the reformulation label for such alignments with a specific category.
    \item ``Addition'': Highlight cases where phrases are present in one span that do not have a direct counterpart in the other segment. Mark these as addition alignments with a specific category.
\end{enumerate}

\subsection{Word-level alignments}
Within each pair of aligned segments (so you cannot create word-level alignment between words that belong to different phrase alignments) labeled translation or paraphrase, you will be annotating word-level alignments, distinguishing between ``sure'' links (direct translations) and ``possible'' links (including additional contextual information, determiners, cases, etc.).
\begin{itemize}
    \item Sure Links (Direct Translation):
Identify and mark word alignments that represent direct translations without any additional context.
These alignments should reflect one-to-one correspondence between words with good translation equivalence.
\item Possible Links (Additional Context):
Identify and mark word alignments where additional contextual information or linguistic elements (such as determiners, cases, etc.) are present in one language and not in the other.
These alignments are not for cross-language counterparts but indicate related, supplementary, or partial information.
\end{itemize}

\end{document}